\documentclass[journal]{IEEEtran}
\usepackage{amsmath,graphicx,cite,epsfig}
\usepackage{multirow}
\usepackage{booktabs}
\usepackage{float}
\usepackage{subfigure}
\usepackage{caption}
\usepackage{threeparttable}
\usepackage{diagbox}
\usepackage{url}
\usepackage{bm}
\usepackage{stfloats}
\UseRawInputEncoding

\ifCLASSINFOpdf

\else

\fi

\begin{document}
%
\title{No-Reference Quality Assessment for 360-degree Images by Analysis of Multi-frequency Information and Local-global Naturalness}
%
%
%

\author{Wei Zhou,~\IEEEmembership{Student~Member,~IEEE}, Jiahua Xu, Qiuping Jiang,~\IEEEmembership{Member,~IEEE}, \\and Zhibo Chen,~\IEEEmembership{Senior~Member,~IEEE}
\thanks{This work was supported in part by NSFC under Grant U1908209, 61632001, 61901236 and the National Key Research and Development Program of China 2018AAA0101400.}
\thanks{W. Zhou, J. Xu and Z. Chen are with the CAS Key Laboratory of Technology in Geo-Spatial Information Processing and Application System, University of Science and Technology of China, Hefei 230027, China (e-mail: weichou@mail.ustc.edu.cn; xujiahua@mail.ustc.edu.cn; chenzhibo@ustc.edu.cn). Q. Jiang is with the School of Information Science and Engineering, Ningbo University, Ningbo 315211, China (e-mail: jiangqiuping@nbu.edu.cn).}
\thanks{}}

\maketitle

\begin{abstract}
360-degree/omnidirectional images (OIs) have achieved remarkable attentions due to the increasing applications of virtual reality (VR). Compared to conventional 2D images, OIs can provide more immersive experience to consumers, benefitting from the higher resolution and plentiful field of views (FoVs). Moreover, observing OIs is usually in the head mounted display (HMD) without references. Therefore, an efficient blind quality assessment method, which is specifically designed for 360-degree images, is urgently desired. In this paper, motivated by the characteristics of the human visual system (HVS) and the viewing process of VR visual contents, we propose a novel and effective no-reference omnidirectional image quality assessment (NR OIQA) algorithm by Multi-Frequency Information and Local-Global Naturalness (MFILGN). Specifically, inspired by the frequency-dependent property of visual cortex, we first decompose the projected equirectangular projection (ERP) maps into wavelet subbands by using discrete Haar wavelet transform (DHWT). Then, the entropy intensities of low-frequency and high-frequency subbands are exploited to measure the multi-frequency information of OIs. Besides, except for considering the global naturalness of ERP maps, owing to the browsed FoVs, we extract the natural scene statistics (NSS) features from each viewport image as the measure of local naturalness. With the proposed multi-frequency information measurement and local-global naturalness measurement, we utilize support vector regression (SVR) as the final image quality regressor to train the quality evaluation model from visual quality-related features to human ratings. To our knowledge, the proposed model is the first no-reference quality assessment method for 360-degreee images that combines multi-frequency information and image naturalness. Experimental results on two publicly available OIQA databases demonstrate that our proposed MFILGN outperforms state-of-the-art full-reference (FR) and NR approaches.
\end{abstract}


%
\IEEEpeerreviewmaketitle

\section{Introduction}
%
%
%
%

\IEEEPARstart{I}{mmersive} multimedia technologies, especially the virtual reality (VR), can provide viewers with more realistic and interactive user experience \cite{diemer2015impact}. As the most common form of VR contents, 360-degree/omnidirectional images and videos record visual information that covers the entire $180\times {{360}^{{}^\circ }}$ viewing spherical, thus attract a lot of attentions from both academy and industry during recent years \cite{xu2020state}. With the commercial head mount display (HMD), users are allowed to freely view any direction with the specific content by head movement, which is different from conventional 2D images and videos. Moreover, due to the omnidirectional viewing range, the resolution of 360-degree image/video is usually ultra-high, e.g. 4K, 8K, or even higher. These would bring many difficulties to the whole 360-degree image/video processing chain, such as acquisition, compression, transmission, reconstruction and display, etc \cite{zhou2018stereoscopic}. Additionally, the perceptual quality of omnidirectional images could be degraded in the 360-degree image/video processing systems. Therefore, the study on 360-degree/omnidirectional image quality assessment (OIQA) is more challenging and significant to guide the development of VR applications.

Recently, there have witnessed ever-increasing interests in the research field of image quality assessment (IQA). Two types of IQA methods are involved, which consist of subjective IQA \cite{zhou20163d,zhou2018visual,xu2018subjective,shi2018perceptual,zhao2019you} and objective IQA \cite{chen2017blind,jiang2017optimizing,jiang2018unified,xu2020binocular,jiang2020full}. In subjective IQA tests, subjects are asked to give the human ratings for each viewed image. After data processing and outlier elimination, the mean opinion score (MOS) can be obtained by computing the average quality scores of all subjects for each image, which can be regarded as the quantitative ground truth of the perceptual quality \cite{zhou2020blind}. Since human is the ultimate viewer, subjective IQA is the most reliable quality assessment approach. By conducting such subjective experiments, several subjective OIQA databases have been established. For example, a testbed for the subjective measurement of 360-degree/omnidirectional contents was proposed \cite{upenik2016testbed}, where 6 reference and 54 distorted omnidirectional images (OIs) were included. The JPEG compression with various quality parameters and two projection models are considered in this database. Moreover, in \cite{upenik2017performance}, 4 high fidelity uncompressed OIs were used to generate 100 impaired images with different geometric projections and three codecs, i.e. JPEG, JPEG2000, and HEVC intra. Four target bitrates were chosen to compress original OIs. The compressed VR image quality database (CVIQD) was built in \cite{sun2017cviqd}, which includes 5 reference and 165 compressed OIs by JPEG, AVC, and HEVC codecs. Furthermore, this database was expanded as the CVIQD2018 \cite{sun2018large} with more visual contents, leading to 16 pristine and 528 compressed images. In \cite{huang2018modeling}, different resolutions and JPEG compression were considered, where the database has 12 original and 144 distorted OIs. In addition, an omnidirectional image quality assessment (OIQA) database \cite{duan2018perceptual} was developed, consisting of 16 reference and 320 distorted images. Apart from JPEG and JPEG2000 compression artifacts, different levels of white Gaussian noise and Gaussian blur were taken into account in this database. Since observers usually exploit the HMD to view OIs, the absolute category rating with hidden reference (ACR-HR) methods, which are also referred to as the single-stimulus (SS) methods, are adopted to build all these subjective quality assessment databases for OIs.

\begin{figure}[t]
	\centerline{\includegraphics[width=9cm]{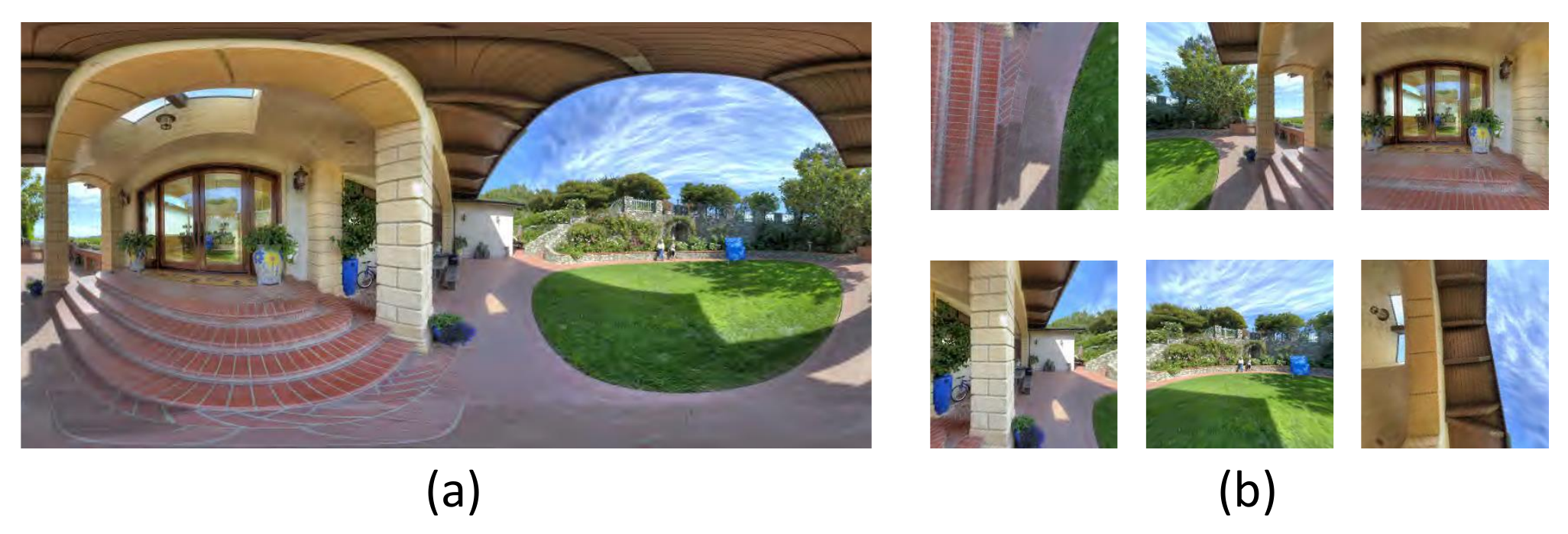}}
	\caption{Illustration of the projected ERP map and various viewports of 360-degree image.}
	\centering
	\label{fig:fig1}
\end{figure}

However, subjective tests are generally time-consuming and labor-intensive. Thus, objective visual quality assessment algorithms used to automatically measure the perceptual quality of OIs are required. As shown in Fig. \ref{fig:fig1}, (a) represents the projected equirectangular projection (ERP) map of a viewed VR scene. The projected ERP map usually has a higher resolution, such as 4K or 8K. Hence, it is suitable for multi-resolution decomposition, which is also an image decomposition in the frequency channels of constant bandwidth on a logarithmic scale. Moreover, the concept of multi-resolution can interpret multi-frequency channel decompositions \cite{mallat1989multifrequency}. In this figure, except for the projected ERP map, (b) shows 6 viewports from a variety of viewing directions, each of which is a part of 360-degree image falling into the field of view (FoV) in the HMD. The naturalness characteristics reflected by statistical regularizations from the global ERP map and local viewports are different. In addition, a no-reference quality assessment model for 3D/stereoscopic omnidirectional images has been proposed in \cite{yang2020latitude}. The monocular multi-scale features are extracted from left and right views separately, while the binocular perception features are exploited from tensor decomposition and the absolute difference map as well as product image of left and right views. The naturalness features involved in the product image are also validated. Nevertheless, it is different from 2D omnidirectional images considered in this paper, where we only have one single view image with multiple viewports.

Based on these observations, in this paper, we propose a novel no-reference (NR) OIQA method by Multi-Frequency Information and Local-Global Naturalness (MFILGN). First, according to a series of studies in neuroscience on the human visual system (HVS), the neuronal responses in the visual cortex are frequency-dependent \cite{heitger1992simulation}. In other words, each neuron corresponds to specific spatial and temporal frequency signals. Therefore, the input visual signal can be decomposed into multiple frequency domains, which is more in line with the human visual perception. Among various multi-frequency channel decomposition methods, wavelet decomposition shows the superiority of processing visual signals \cite{mallat1989theory}. Meanwhile, the wavelet decomposition also has been demonstrated to have good performance in IQA \cite{vu2012fast,wang2019blind}. Motivated by this mechanism, we decompose the projected ERP maps into wavelet subbands through discrete Haar wavelet transform (DHWT). Since the decomposed low-frequency and high-frequency subbands represent luminance information and textural details, respectively. We then compute the entropy intensities of low-frequency and high-frequency subbands, which are used to measure the multi-frequency information of OIs. Second, due to different viewports during browsing, we propose the local-global naturalness measurement, where the natural scene statistics (NSS) features are extracted from both local viewed FoVs and global ERP maps. To the best of our knowledge, the naturalness has been proved in a number of IQA research \cite{shi2019no,zhou2020tensor} but it has not been used in OIQA. Finally, the quality predictions of OIs are obtained by the well-known support vector regression (SVR). As demonstrated by extensive experiments, our proposed MFILGN performs better than state-of-the-art FR and NR algorithms on two publicly available OIQA databases.

\begin{figure*}[t]
	\centerline{\includegraphics[width=18.5cm]{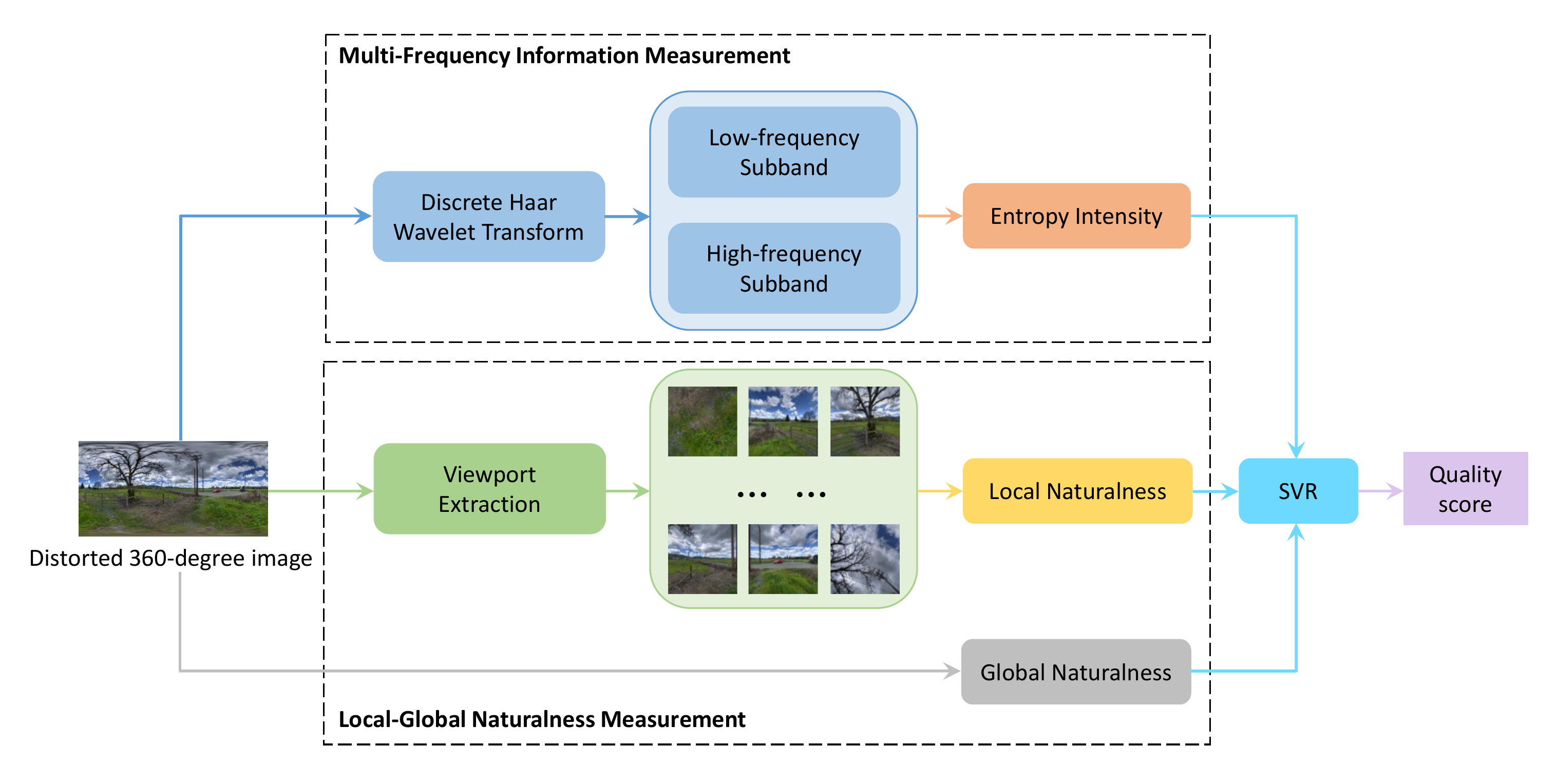}}
	\caption{The overall framework of our proposed MFILGN.  It consists of the multi-frequency information measurement and local-global naturalness measurement.}
	\centering
	\label{fig:fig2}
\end{figure*}

The main contributions of this work are summarized as follows:

\begin{itemize}
\item We propose the first blind OIQA algorithm based on the measurements of multi-frequency information and local-global naturalness.
\item According to the frequency-dependent characteristic of human visual cortex, we derive the decomposed low and high frequency subbands and utilize the entropy intensities of these subband images to reflect the multi-frequency information in omnidirectional images.
\item Considering the viewing process of omnidirectional images, apart from the global naturalness from the projected ERP maps, we extract the local naturalness features from various viewports together with global NSS to composite the local-global naturalness measurement.
\end{itemize}

The remaining sections of this paper are organized as follows. In Section II, we present the related works of objective quality assessment for both traditional 2D images and OIs. Section III introduces the proposed MFILGN model for NR OIQA in details. In Section IV, experimental results and analysis are presented. Section V concludes the whole paper with possible research directions in the future.

\section{Related Works}
Whether for traditional objective IQA or objective OIQA, when the originally pristine/reference image is available, full-reference (FR) objective visual quality assessment models are developed. As for conventional FR IQA, the signal fidelity metrics such as mean square error (MSE) and peak signal-to-noise ratio (PSNR), measure the visual quality by computing the pixel differences between original and distorted images. Due to the simple calculation and optimization processes, they are widely used in image processing. Nevertheless, their performance is relatively unsatisfactory and cannot predict the human perceived visual quality precisely. Therefore, the characteristics of the HVS are employed to construct perception-based IQA models. For instance, the structural similarity (SSIM) index \cite{wang2004image} and its different variants, including the multiscale SSIM (MS-SSIM) index \cite{wang2003multiscale}, the feature similarity (FSIM) index \cite{zhang2011fsim}, etc.

When coming to practical scenarios, perfect-quality original images are usually difficult to obtain, and thus no-reference objective visual quality assessment approaches are urgently needed. For traditional NR IQA, many methods extract hand-crafted distortion-discriminative features for predicting the perceptual image quality, such as the blind/referenceless image spatial quality evaluator (BRISQUE) \cite{mittal2012no}, the blind multiple pseudo reference images-based (BMPRI) measure \cite{min2018blind}, and so on. In recent decades, due to the powerful feature representation learning capacity, deep learning has shown unprecedented success in many image processing and computer vision tasks \cite{jin2018decomposed,zhou2019dual}. This also presents an opportunity for evaluating the image visual quality. Typical methods contain the deep image quality assessment (DeepQA) \cite{kim2017deep} and the deep bilinear convolutional neural network (DBCNN) \cite{zhang2018blind} for FR and NR IQA, respectively. Furthermore, several perception-based pre-processing strategies have been presented in existing works. For example, before training CNN models, saliency maps can be used to assign importance to distorted image patches \cite{jia2018saliency}. The distorted image stream and gradient image stream were both considered in CNN to predict perceptual image quality scores \cite{yan2018two}.

Although there exist lots of classical objective quality assessment algorithms for FR and NR IQA, they are designed for regular flat 2D images and unsuitable for assessing the perceptual quality of OIs. Yet, limited research works related to objective OIQA methods have been proposed in the literature. In general, existing objective OIQA models can be classified into two categories. The first is to extend conventional FR IQA approaches to FR OIQA \cite{yu2015framework,sun2017weighted,zakharchenko2016quality,xu2018assessing,zhou2018weighted,chen2018spherical}. For example, several PSNR-based FR OIQA models were presented. Yu \textit{et al.} \cite{yu2015framework} proposed the spherical PSNR (S-PSNR), which chose specific points on the spherical surface rather than the projected panoramic image. Sun \textit{et al.} \cite{sun2017weighted} developed the weighted-to-spherically-uniform PSNR (WS-PSNR) by combining the error map with the weighted map which was determined by stretched regions. Zakharchenko \textit{et al.} \cite{zakharchenko2016quality} presented the craster parabolic projection PSNR (CPP-PSNR), which computed the PSNR on the craster parabolic projection domain. Xu \textit{et al.} \cite{xu2018assessing} put forward the non-content-based PSNR (NCP-PSNR) and content-based PSNR (CP-PSNR) via weighting pixels with position information and predicting viewing direction, respectively. Likewise, the performance of PSNR-based FR OIQA methods is insufficient because they do not consider the HVS characteristics. Afterwards, objective FR OIQA methods based on SSIM have been proposed in succession. Similar to WS-PSNR, Zhou \textit{et al.} \cite{zhou2018weighted} designed the weighted-to-spherically-uniform SSIM (WS-SSIM), where the position weighted map was used to multiply the SSIM. In order to reduce the influence of geometric distortions for the projection, Chen \textit{et al.} \cite{chen2018spherical} proposed the spherical SSIM (S-SSIM), which calculated the similarity of each image pixel on the sphere. The second is the deep learning-based NR IQA methods \cite{kim2019deep,li2018bridge,li2019viewport,sun2019mc360iqa,xu2020blind}. Considering the spherical representation of 360-degree content, Kim \textit{et al.} \cite{kim2019deep} proposed a deep learning framework for VR image quality assessment (DeepVR-IQA) based on adversarial learning, in which the quality scores of sampled patches were predicted and then they were weighted according to the patch positions on the sphere. Moreover, in \cite{li2018bridge}, the head movement (HM) and eye movement (EM) were exploited to weight the quality scores in deep learning models. But the image patches sampled from the projected plane reveal non-negligible geometric deformation, which cannot reflect the actual viewing contents. Thus, viewports were utilized to predict the perceptual quality of OIs. Specifically, Li \textit{et al.} \cite{li2019viewport} proposed a viewport-based CNN (V-CNN), which predicted the quality scores of viewports instead of image patches sampled from projected planes. Sun \textit{et al.} \cite{sun2019mc360iqa} designed a multi-channel CNN for blind 360-degree image quality assessment (MC360IQA), including 6 parallel hyper-ResNet34 networks to process viewport images and an image quality regression module to aggregate learned features for obtaining the final image quality. Xu \textit{et al.} \cite{xu2020blind} proposed the viewport oriented graph convolution network (VGCN) to tackle the perceptual quality assessment for OIs, which was guided by different viewports.

From the above reviewed objective quality assessment models for OIs, it can be concluded that existing objective OIQA methods have achieved success to a certain extent. However, we also notice that they all ignore the important multi-frequency information as well as the statistical regularizations of OIs from both global projection maps and local viewports. To fill the blank, in this work, we try to present a no-reference OIQA method by considering multi-frequency information and local-global naturalness simultaneously.

\section{The Proposed Quality Assessment Method}
In this section, we will introduce the proposed NR OIQA method that can blindly predict the perceptual quality of OIs in technical details. The overall framework of our proposed MFILGN is shown in Fig. \ref{fig:fig2}, which consists of two separate measurements, namely the multi-frequency information measurement and local-global naturalness measurement. For the multi-frequency information measurement, inspired by the HVS characteristics, the Haar wavelet transform is first applied to decompose the projected ERP maps into multiple subbands. Then, the entropy intensities are calculated for measuring the multi-frequency information. And for the local-global naturalness measurement, considering the changeable FoVs during the viewing process, both local naturalness from different viewports and global naturalness from ERP maps are extracted. The final quality index is obtained by the regression of these distortion-related features.

\subsection{Image Decomposition}
\begin{figure}[t]
	\centerline{\includegraphics[width=9cm]{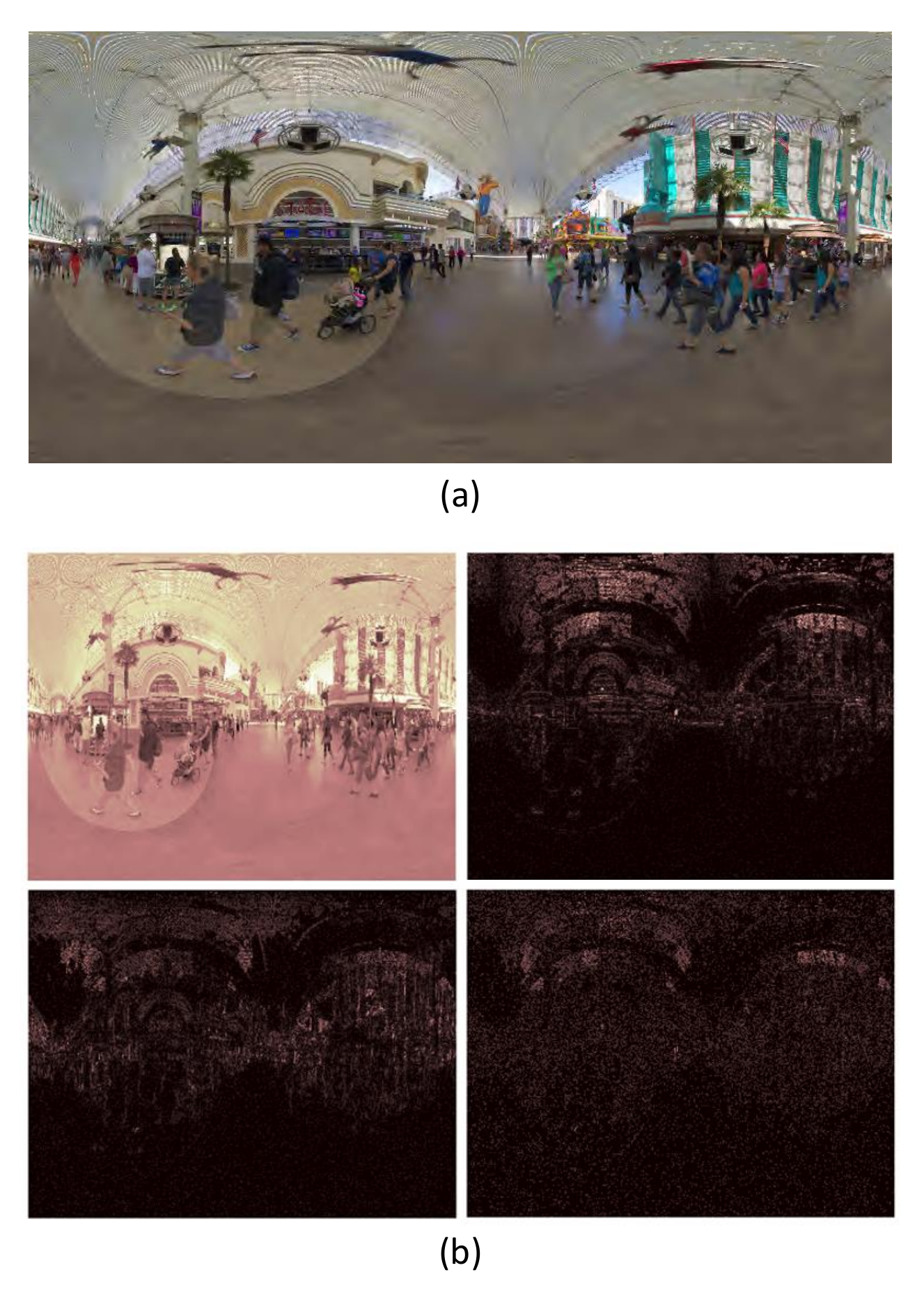}}
	\caption{Multi-frequency channel decomposition by DHWT. (a) An example of distorted 360-degree image; (b) the decomposed four subband images of (a).}
	\centering
	\label{fig:fig3}
\end{figure}

\begin{figure*}[t]
	\centerline{\includegraphics[width=14cm]{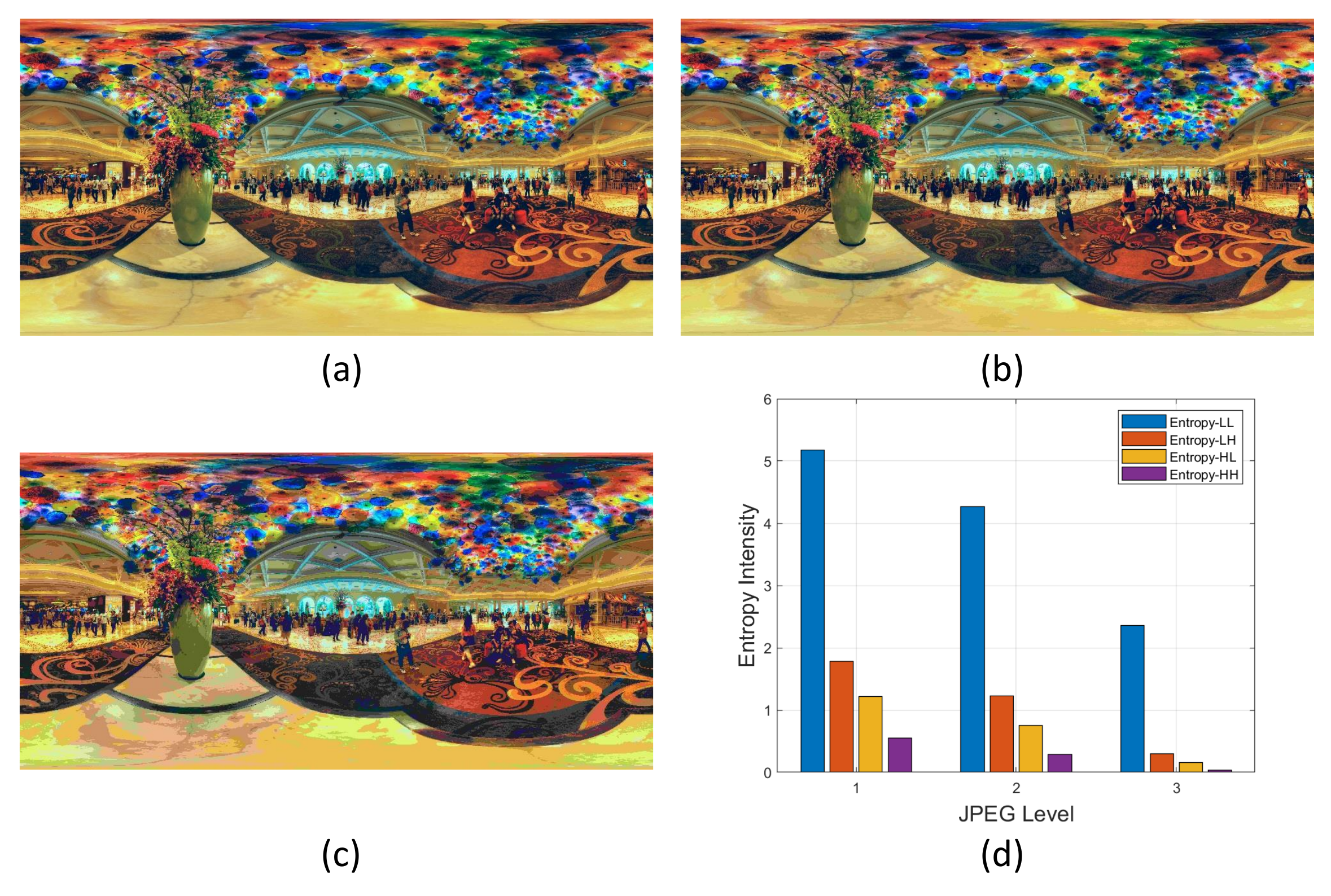}}
	\caption{Entropy intensity changes for different distorted 360-images. (a) Low JPEG distortion denoted by JPEG-Level1; (b) Middle JPEG distortion denoted by JPEG-Level2; (c) High JPEG distortion denoted by JPEG-Level3; (d) The corresponding entropy intensity changes of low and high frequency subbands for (a), (b) and (c).}
	\centering
	\label{fig:fig4}
\end{figure*}

According to the HVS studies, different neurons respond to different frequencies of visual signals \cite{heitger1992simulation}. The input visual stimulus should be decomposed into various subband images for the subsequent processing. Moreover, wavelet transform is one of the multi-frequency channel decomposition methods.  Here, we choose the wavelet transform to conduct the distorted image decomposition. Generally, there exist a number of wavelets, including Haar wavelet, Morse wavelet, Gabor wavelet, Bump wavelet and so on. In all of these wavelets, Haar wavelet is symmetric and a special case of Daubechies wavelet, which shows great success in perceptual quality assessment \cite{lai2000haar,reisenhofer2018haar,zhu2019multi}. Therefore, we adopt the DHWT to decompose the distorted OIs into multi-frequency subbands. Specifically, the Haar wavelet can be formulated as:

\begin{equation}\label{1}
\begin{aligned}
\frac{1}{\sqrt{2}}\psi (t)&=\frac{1}{\sqrt{2}}(\phi (t-1)-\phi (t))\\
&=\sum\limits_{u=-\infty }^{+\infty }{{{(-1)}^{1-u}}h[1-u]}\phi (t-u),
\end{aligned}
\end{equation}
where the mother wavelet $\psi$ is defined by:

\begin{equation}\label{2}
\psi (t)=\left\{
\begin{aligned}
1,\text{   }for\text{ }t\in [0,\text{ }\frac{1}{2}), \\
-1,\text{ }for\text{ }t\in [\frac{1}{2},\text{ 1}), \\
0,\text{ }otherwise.
\end{aligned}
\right.
\end{equation}
Moreover, the father wavelet or scaling function $\phi$ and its corresponding filter $h$ are computed as:

\begin{equation}\label{3}
\phi (t)=\left\{
\begin{aligned}
1,\text{   }for\text{ }t\in [0,\text{ 1}), \\
0,\text{ }otherwise.
\end{aligned}
\right.
\end{equation}

\begin{equation}\label{4}
h[u]=\left\{
\begin{aligned}
\frac{1}{\sqrt{2}},\text{  }for\text{ }u=0,1, \\
0,\text{     }otherwise.
\end{aligned}
\right.
\end{equation}

With the Haar wavelet, we can obtain the Haar wavelet transform by cross multiplying different shifts and stretches. Here, let $H$ denote the DHWT matrix. Suppose that a distorted 360-degree image $D$ with resolution $I\times J$. Then the input 360-degree image can be decomposed into four $\frac{I}{2}\times \frac{J}{2}$ subband images by the DHWT as follows:

\begin{equation}\label{5}
HD{{H}^{T}}=\left[ \begin{matrix}
   {{D}_{LL}} & {{D}_{HL}}  \\
   {{D}_{LH}} & {{D}_{HH}}  \\
\end{matrix} \right],
\end{equation}
where ${D}_{LL}$, ${D}_{HL}$, ${D}_{LH}$, and ${D}_{HH}$ represent the decomposed four subbands with low or high frequency in the horizontal or vertical direction. ${H}^{T}$ is the transpose matrix of $H$.

\subsection{Multi-frequency Information}
\begin{figure*}[t]
	\centerline{\includegraphics[width=18cm]{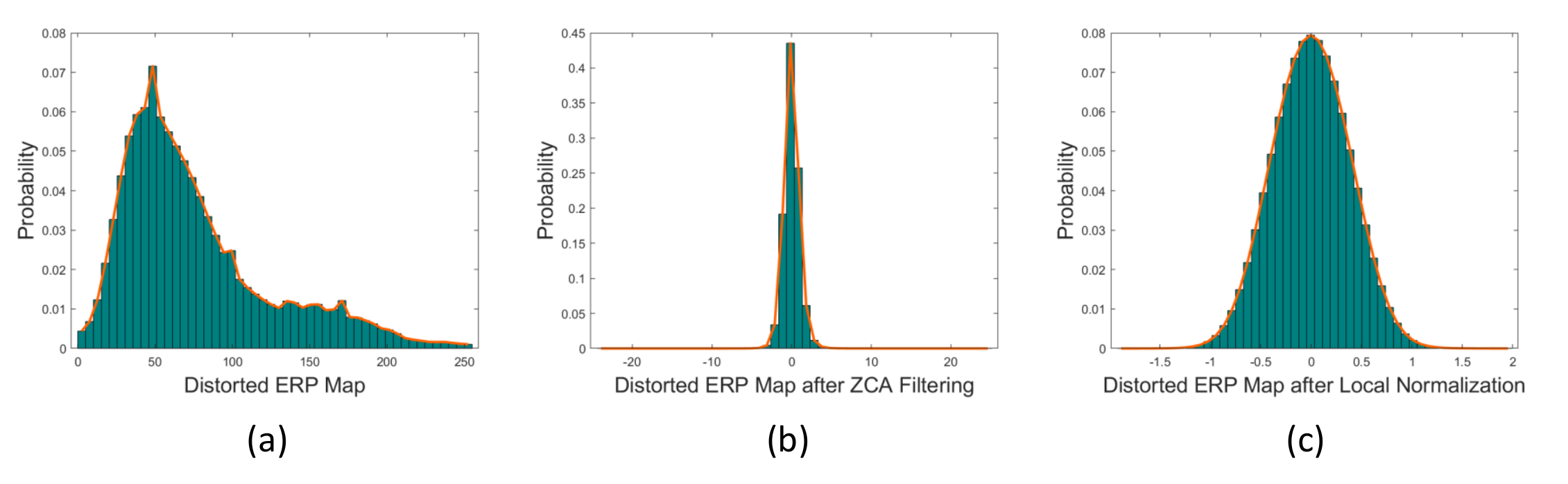}}
	\caption{Effects of the ZCA whitening filter and MSCN operation on the statistical distribution of distorted ERP map.}
	\centering
	\label{fig:fig5}
\end{figure*}

Fig. \ref{fig:fig3} illustrates the multi-frequency channel decomposition by DHWT. We show a distorted 360-degree image in (a), and the four images in (b) are the decomposed results of (a). From this figure, we can see that the decomposed subband images are different from each other, which could reflect various frequency characteristics of the distorted 360-degree image. Since the entropy intensities could reflect average amount of image information, images with different distortions bring about various entropy intensities. Especially, after the multi-frequency channel decompositions of distorted 360-degree images, the entropy intensities of decomposed low-frequency and high-frequency subbands differ from each other. We then compute the entropy intensities of the decomposed four subband images as:

\begin{equation}\label{6}
{{E}_{s}}=-\sum\limits_{k=0}^{K}{p_{k}^{s}{{\log }_{2}}p_{k}^{s}},
\end{equation}
where $s\in \{LL,\text{ }HL,\text{ }LH,\text{ }HH\}$ denotes the collection of frequency decomposition components. $p_{k}^{s}$ and $K$ indicate the probability of the pixel equaling to $k$ and the maximum pixel value in the corresponding decomposed subband image. The probability can be calculated by:

\begin{equation}\label{7}
p_{k}^{s}=\frac{N_{k}^{s}}{N},
\end{equation}
where $N_{k}^{s}$ is the numbers of pixel value equaling to $k$ in the corresponding decomposed subband image. $N$ represents the total number of pixels.

After computing the entropy intensities of the decomposed four subband images, we use their joint component features to measure multi-frequency information as:

\begin{equation}\label{8}
{{F}_{MFI}}=[{{E}_{LL}},\text{ }{{E}_{HL}},\text{ }{{E}_{LH}},\text{ }{{E}_{HH}}],
\end{equation}
where $F_{MFI}$ denotes the multi-frequency information measurement, which reflects the discriminative entropy information from both low and high frequency subbands.

In Fig. \ref{fig:fig4}, we show three 360-degree images with different JPEG distortion levels and their corresponding entropy intensities of low-frequency and high-frequency subbands. The image samples are from OIQA database \cite{duan2018perceptual}. We can find that the entropy intensities for four frequency subbands change significantly with regard to various distortions. That is, more JPEG distortions lead to a decrease in entropy intensities of low and high frequency subbands, which demonstrates the effectiveness of the proposed multi-frequency information measurement.

\subsection{Global Naturalness}
In addition to the multi-frequency information in the projected ERP maps, the image naturalness which is reflected by statistical regularizations is crucial to the perceptual quality of OIs. Although the NSS features have been applied to traditional multimedia formats, such as 2D \cite{mittal2012no}, 3D \cite{chen2017blind}, etc. To the best of our knowledge, it has not been used to evaluate the perceptual quality of VR images yet. Therefore, here we try to explore the naturalness in OIQA. Intuitively, we can extract the NSS features from global ERP maps. Given the distorted 360-degree image $D$ with resolution $I\times J$, to reduce the spatial redundancy of adjacent image pixels, we first adopt the zero-phase component analysis (ZCA) whitening filter as follows:

\begin{equation}\label{9}
{{D}^{z}}=Z(D),
\end{equation}
where $Z$ indicates the ZCA whitening filter. ${D}^{z}$ is the distorted ERP map after ZCA filtering.

Afterwards, the local mean subtracted and contrast normalized (MSCN) coefficients are computed to measure the image naturalness, which can model the contrast gain masking in early human visual cortex \cite{carandini1997linearity}. For each distorted ERP map after ZCA filtering, the MSCN coefficients are calculated by:

\begin{equation}\label{10}
{{\hat{D}}^{z}}(i,j)=\frac{{{D}^{z}}(i,j)-\mu (i,j)}{\sigma (i,j)+C},
\end{equation}
where ${{\hat{D}}^{z}}(i,j)$ and ${{D}^{z}}(i,j)$ are the distorted ERP map after local normalization (i.e. the MSCN coefficients) and distorted ERP map at spatial position $(i,j)$, respectively. $\mu (i,j)$ and $\sigma (i,j)$ represent the local mean and standard deviation of the distorted ERP map, which are computed as:

\begin{equation}\label{11}
\mu (i,j)=\sum\limits_{s=-S}^{S}{\sum\limits_{t=-T}^{T}{{{w}_{s,t}}}}{{D}^{z}}(i,j),
\end{equation}

\begin{equation}\label{12}
\sigma (i,j)=\sqrt{\sum\limits_{s=-S}^{S}{\sum\limits_{t=-T}^{T}{{{w}_{s,t}}(}}{{D}^{z}}(i,j)-\mu (i,j){{)}^{2}}},
\end{equation}
where $w=\{{{w}_{s,t}}|s=-S,...,S,t=-T,...T\}$ stands for the 2D circularly-symmetric Gaussian weighted function.

\begin{figure*}[t]
  \centerline{\includegraphics[width=16cm]{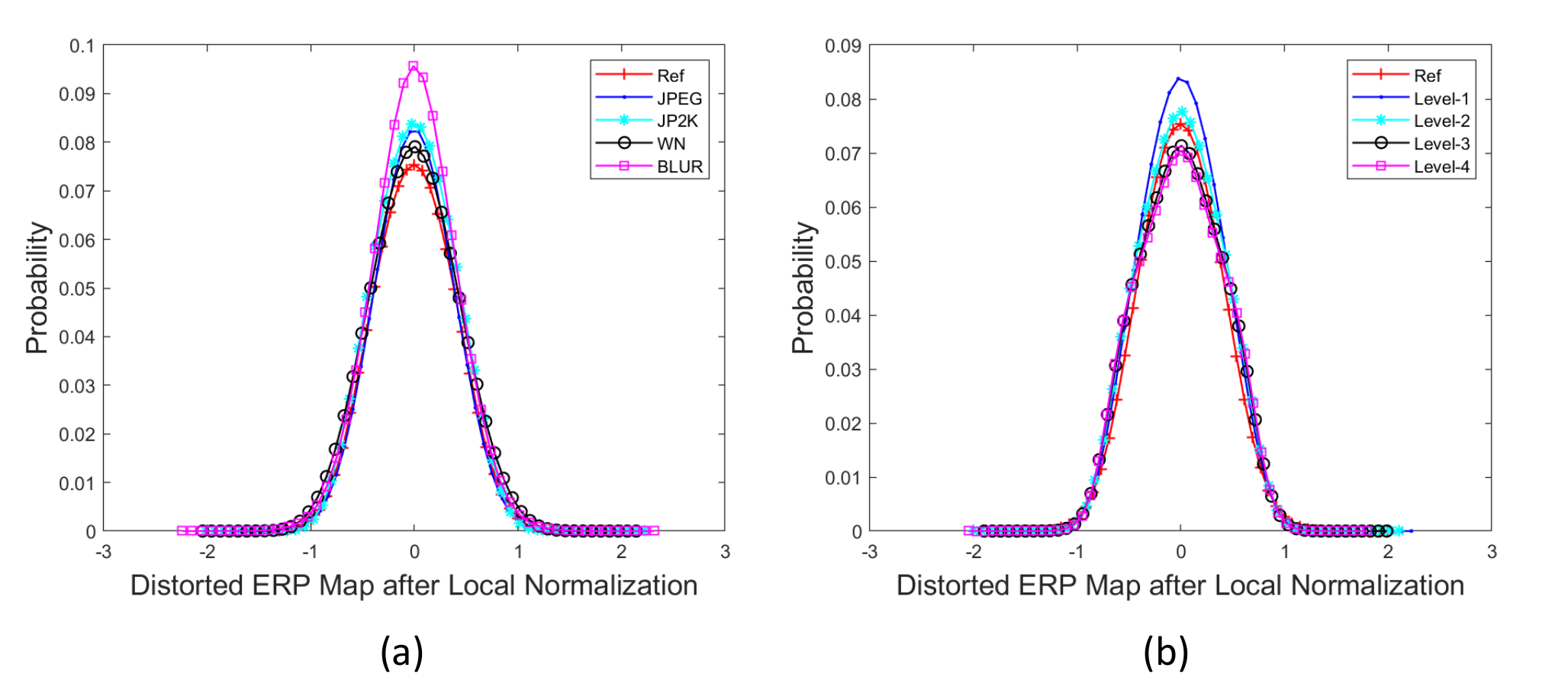}}
  \caption{Statistical distributions of MSCN coefficients for different distortion types and levels from OIQA database \cite{duan2018perceptual}. (a) MSCN coefficients vary with various distortion types; (b) MSCN coefficients under four JPEG2000 compression levels.}
  \centering
\label{fig:fig6}
\end{figure*}

In Fig. \ref{fig:fig5}, we show the statistical distributions of input distorted ERP map and the distorted ERP map after ZCA filtering as well as local normalization process. From figures (a-c), we can observe that the ZCA filtering and the MSCN operation both make the probability distribution of the distorted ERP map more Gaussian-like. Moreover, the statistical distribution after local normalization is the closest to Gaussian distribution. Thus, we exploit the MSCN coefficients for the subsequent feature processing. Additionally, Fig. \ref{fig:fig6} presents the statistical distributions of MSCN coefficients for different distortion types and levels. As can be seen in this figure, the probability distributions of the distorted ERP map after local normalization is influenced by different distortion types as well as the distortion levels, which demonstrates that the statistical distributions of MSCN coefficients are discriminative for the perceptual quality assessment of 360-degree images.

With the probability distributions of the distorted ERP map after local normalization, we utilize the zero-mean generalized Gaussian distribution (GGD) and asymmetric generalized Gaussian distribution (AGGD) models to quantify the MSCN coefficients distribution. The zero-mean AGGD used to fit the distribution is:

\begin{equation}\label{13}
f(x;\tau ,\sigma _{l}^{2},\sigma _{r}^{2})=\left\{
\begin{aligned}
\frac{\tau }{({{\upsilon }_{l}}+{{\upsilon }_{r}})\Gamma (\frac{1}{\tau })}{{e}^{-{{(\frac{-x}{{{\upsilon }_{l}}})}^{\tau }}}},\text{ }x<0, \\
\frac{\tau }{({{\upsilon }_{l}}+{{\upsilon }_{r}})\Gamma (\frac{1}{\tau })}{{e}^{-{{(\frac{-x}{{{\upsilon }_{r}}})}^{\tau }}}},\text{ }x\ge 0,
\end{aligned}
\right.
\end{equation}
where

\begin{equation}\label{14}
{{\upsilon }_{l}}={{\sigma }_{l}}\sqrt{\frac{\Gamma (\frac{1}{\tau })}{\Gamma (\frac{3}{\tau })}},
\end{equation}

\begin{equation}\label{15}
{{\upsilon }_{r}}={{\sigma }_{r}}\sqrt{\frac{\Gamma (\frac{1}{\tau })}{\Gamma (\frac{3}{\tau })}},
\end{equation}
and $\tau$ denotes the shape parameter which can control the shape of the statistical distribution. ${\sigma }_{l}$ and ${\sigma }_{r}$ are the scales of the left and right sides for the statistical distribution, respectively. $\Gamma(\cdot)$ is the gamma function that is defined as:

\begin{equation}\label{16}
\Gamma (a)=\int\limits_{0}^{+\infty }{{{x}^{a-1}}}{{e}^{-x}}dx,\text{ }a>0.
\end{equation}
The best AGGD fitting with parameters $(\eta ,\tau ,\sigma _{l}^{2},\sigma _{r}^{2})$ are then calculated and $\eta$ is given by:

\begin{equation}\label{17}
\eta =({{\upsilon }_{l}}-{{\upsilon }_{r}})\frac{\Gamma (\frac{2}{\tau })}{\Gamma (\frac{1}{\tau })}.
\end{equation}
Furthermore, if ${{\sigma }_{l}}={{\sigma }_{r}}$, the AGGD becomes GGD model as follows:

\begin{equation}\label{18}
f(x;\tau ,{{\sigma }^{2}})=\frac{\tau }{2\upsilon \Gamma (\frac{1}{\tau })}{{e}^{-{{(\frac{|x|}{\upsilon })}^{\tau }}}},
\end{equation}
where

\begin{equation}\label{19}
\upsilon =\sigma \sqrt{\frac{\Gamma (\frac{1}{\tau })}{\Gamma (\frac{3}{\tau })}}.
\end{equation}

In addition, the distorted 360-degree images are downsampled by a factor of 2. Finally, the two scales, including the original image scale and a reduced resolution scale, are exploited to extract the global NSS features $F_{GNSS}$ from the projected ERP maps.

\subsection{Local Naturalness}
\begin{figure*}[t]
  \centerline{\includegraphics[width=18cm]{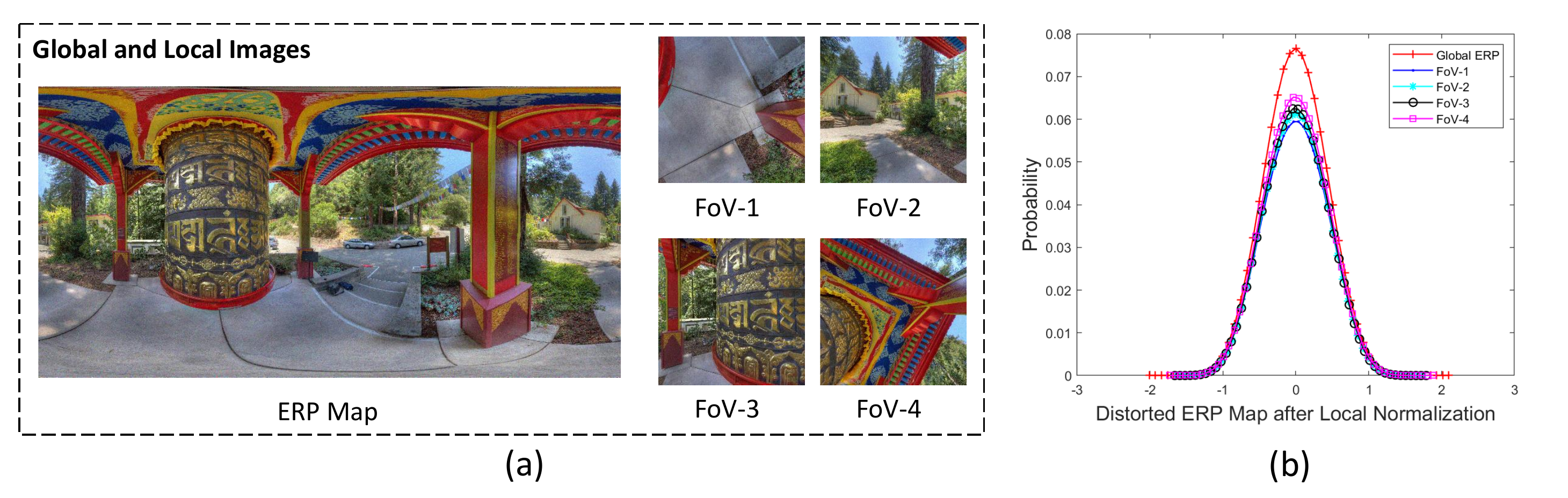}}
  \caption{An example of global and local images from different spatial positions with their corresponding statistical distributions of MSCN coefficients.}
  \centering
\label{fig:fig7}
\end{figure*}

When observers browse VR images in the HMD, they can freely change the viewports. Moreover, different FoVs have various contents and the global 360-degree scenery could be reconstructed based on what human see from multiple viewports. Therefore, apart from the global naturalness from the projected ERP maps, it is also important to explore the local naturalness according to a variety of FoVs. As illustrated in Fig. \ref{fig:fig7} (a), we show an example of global ERP map and local images from different spatial positions (i.e. the four FoVs). (b) is the corresponding statistical distributions of MSCN coefficients. It can be seen that the probability distributions of global ERP map and four FoVs are different from each other. Besides, the distributions of FoV-2 and FoV-3 nearing the equator is close. Thus, we adopt the local NSS from these FoVs located at various spatial positions as the complementary features to global NSS from the projected ERP maps.

Specially, suppose that there exist $M$ FoV images, we extract the NSS features from them by the same steps as global naturalness. We then can obtain $(F_{LNSS}^{1},F_{LNSS}^{2},...,F_{LNSS}^{M})$ representing all these viewports. Finally, the average of the NSS features are computed by:

\begin{equation}\label{20}
{{F}_{LNSS}}=\frac{1}{M}\sum\limits_{m=1}^{M}{F_{LNSS}^{m}},
\end{equation}
where ${F}_{LNSS}$ indicates the local NSS features from viewed FoVs. With the local and global NSS features, we combine them to constitute the local-global naturalness measurement as follows:

\begin{equation}\label{21}
{{F}_{LGN}}=[{{F}_{LNSS}},\text{ }{{F}_{GNSS}}].
\end{equation}

In order to achieve the local-global naturalness measurement, we need to sample $M$ viewports from each distorted 360-degree image. Motivated by that VR images are viewed on a sphere and polar regions usually stretch resulting in geometric deformation for projected ERP maps. Thus, we employ the non-uniform sampling strategy \cite{xu2019quality,chen2020stereoscopic}. As shown in Fig. \ref{fig:fig8}, for the equator, we first sample ${M}_{0}$ viewports equidistantly. Then, the remaining viewpoints are selected by:

\begin{equation}\label{22}
\theta =\frac{{{360}^{\circ }}}{{{M}_{0}}},
\end{equation}

\begin{equation}\label{23}
{{M}_{1}}=\left\lfloor {{M}_{0}}\cos \theta  \right\rfloor,
\end{equation}

\begin{equation}\label{24}
{{M}_{2}}=\left\lfloor {{M}_{0}}\cos 2\theta  \right\rfloor,
\end{equation}

\begin{equation}\label{25}
{{M}_{end}}=\left\lfloor {{M}_{0}}\cos \frac{{{90}^{\circ }}}{\theta } \right\rfloor,
\end{equation}
where ${M}_{1}$ and ${M}_{2}$ are the numbers of viewports sampled on $\theta$ and $2\theta$ degrees north or south latitude, respectively. The sampling process ends when the maximum latitude reaches ${90}^{\circ }$ with ${M}_{end}$ sampled viewports. After the computation of viewports sampling, we can totally obtain $M$ viewports for each distorted OI as:

\begin{equation}\label{26}
M={{M}_{0}}+\sum\limits_{m=1}^{end}{2* {{M}_{m}}}.
\end{equation}

\subsection{Quality Regression}
With the multi-frequency information measurement and local-global naturalness measurement, the ultimate quality score of 360-degree image is obtained by the well-known SVR \cite{scholkopf2000new}. Specifically, we randomly divide all distorted OIs into a training set and a testing set, which are denoted by ${{\chi }_{train}}$ and ${{\chi }_{test}}$, respectively. By adopting the SVR, our proposed MFILGN model is achieved from training the distortion-related features of OIs in the training set ${{\chi }_{train}}$ and their corresponding MOS values. Given a distorted 360-degree image ${{D}_{train}}\in {{\chi }_{train}}$ and its extracted features $[{{F}_{MFI}},{{F}_{LGN}}]$ from the multi-frequency information measurement and local-global naturalness measurement, the MFILGN model is defined by:

\begin{equation}\label{27}
MFILGN=SVR\_TRAIN([{{F}_{MFI}}, {{F}_{LGN}}], [Q]),
\end{equation}
where $Q$ is the subjective quality rating (i.e. MOS) of the input 360-degree image.

After training the proposed MFILGN model, we verify the model performance on the testing set ${{\chi }_{test}}$. For example, the predicted quality score of a tested 360-degree image ${{D}_{test}}\in {{\chi }_{test}}$ that does not appear in the training set is computed as follows:

\begin{equation}\label{28}
q=SVR\_PREDICT([{{\widehat{F}}_{MFI}}, {{\widehat{F}}_{LGN}}], MFILGN),
\end{equation}
where $[{{\widehat{F}}_{MFI}},{{\widehat{F}}_{LGN}}]$ represent the extracted features from the multi-frequency information measurement and local-global naturalness measurement for the tested 360-degree image. Finally, the correlation or error between the predicted scores and the corresponding ground truth MOS values for testing set is measured as the performance of MFILGN.

\section{Experimental Results and Analysis}
In this section, we first introduce the experimental protocol including OIQA databases and measure criteria: used in our experiments. Then, we evaluate the proposed MFILGN for overall performance and performance for individual distortion type on the OIQA \cite{duan2018perceptual} and CVIQD \cite{sun2017cviqd,sun2018large} databases. After that, various weighting methods of viewports as well as different parameters containing the adopted viewport numbers and training percentages are analyzed. Finally, the ablation study is conducted to prove the effectiveness of each component in our MFILGN model.

\begin{figure*}[t]
  \centerline{\includegraphics[width=18cm]{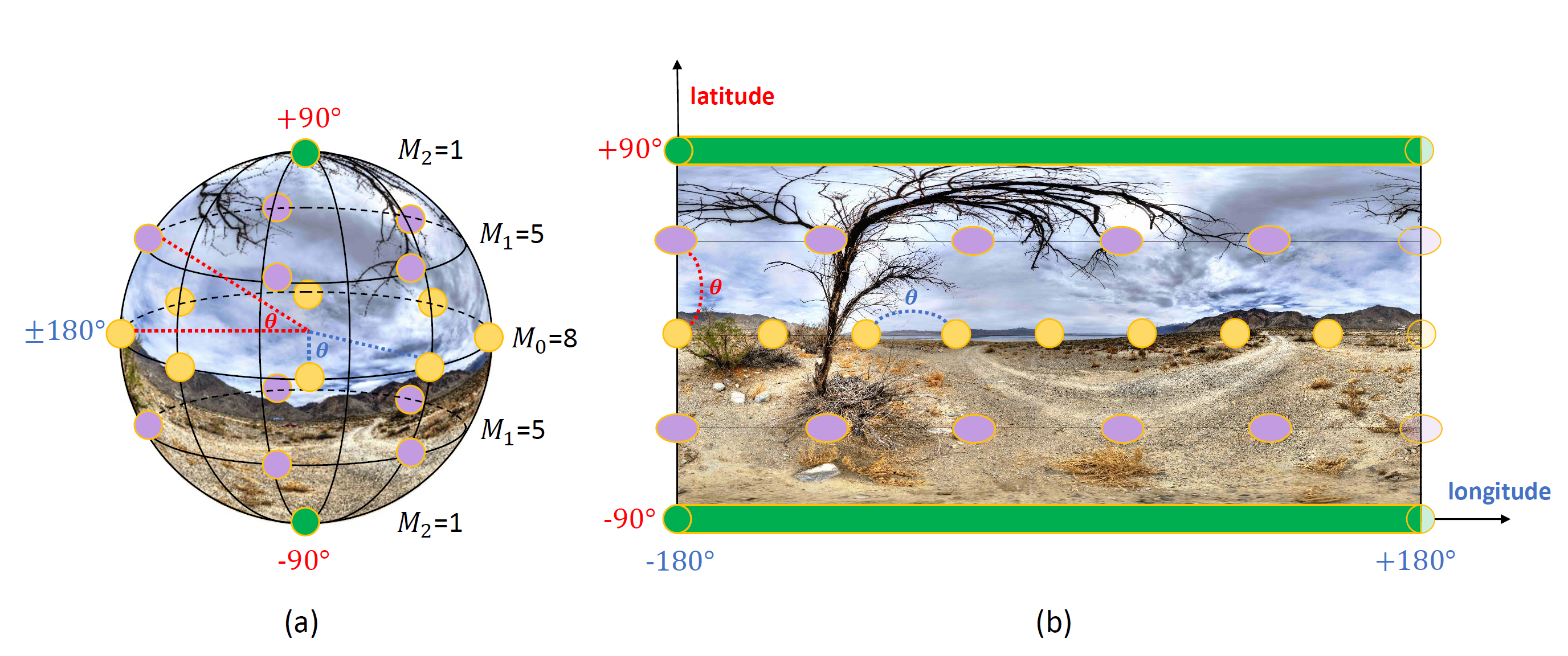}}
  \caption{Demonstration of sampling viewports when ${{M}_{0}}=8$, $\theta ={{45}^{\circ }}$ for the sphere and plane, respectively.}
  \centering
\label{fig:fig8}
\end{figure*}

\begin{table*}[t]
	\begin{center}
		\captionsetup{justification=centering}
		\caption{\textsc{Overall Performance Comparisons on OIQA and CVIQD Databases. The Best Results are denoted in Bold.}}
		\label{table1}
		\scalebox{1.2}{
			\begin{tabular}{c|c|ccc|ccc}
				\toprule
				& Database  & \multicolumn{3}{c|}{OIQA} & \multicolumn{3}{c}{CVIQD} \\ \midrule
				Type & Methods & SROCC & PLCC & RMSE & SROCC & PLCC & RMSE \\ \midrule
				\multirow{8}{*}{FR} & PSNR & 0.5226 & 0.5812 & 1.7005 & 0.6239 & 0.7008 & 9.9599 \\
				& S-PSNR \cite{yu2015framework} & 0.5399 & 0.5997 & 1.6721 & 0.6449 & 0.7083 & 9.8564 \\
				& WS-PSNR \cite{sun2017weighted} & 0.5263 & 0.5819 & 1.6994 & 0.6107 & 0.6729 & 10.3283 \\
				& CPP-PSNR \cite{zakharchenko2016quality} & 0.5149 & 0.5683 & 1.7193 & 0.6265 & 0.6871 & 10.1448 \\
                & SSIM \cite{wang2004image} & 0.8588 & 0.8718 & 1.0238 & 0.8842 & 0.9002 & 6.0793 \\
                & MS-SSIM \cite{wang2003multiscale} & 0.7379 & 0.7710 & 1.3308 & 0.8222 & 0.8521 & 7.3072 \\
                & FSIM \cite{zhang2011fsim} & 0.8938 & 0.9014 & 0.9047 & 0.9152 & 0.9340	& 4.9864 \\
				& DeepQA \cite{kim2017deep} & 0.8973 & 0.9044 & 0.8914 & 0.9292 & 0.9375 & 4.8574 \\ \midrule
				\multirow{6}{*}{NR} & BRISQUE \cite{mittal2012no} & 0.8331 & 0.8424 & 1.1261 & 0.8180 & 0.8376 & 7.6271 \\
				& BMPRI \cite{min2018blind} & 0.6238	& 0.6503 & 1.5874 & 0.7470 & 0.7919	& 8.5258 \\
				& DB-CNN \cite{zhang2018blind} & 0.8653 & 0.8852 &0.9717 & 0.9308 & 0.9356 & 4.9311 \\
				& MC360IQA \cite{sun2019mc360iqa} & 0.9139 & 0.9267 & 0.7854	& 0.9428 & 0.9429 & 4.6506 \\
				& VGCN \cite{xu2020blind} & 0.9515 & 0.9584 & 0.5967 & 0.9639 & 0.9651 & 3.6573 \\
				& Proposed MFILGN & \textbf{0.9614} & \textbf{0.9695} & \textbf{0.5146} & \textbf{0.9670} & \textbf{0.9751} & \textbf{3.1036} \\ \bottomrule
		\end{tabular}}
	\end{center}
\end{table*}

\subsection{Experimental Protocol}
\begin{table*}[t]
	\begin{center}
		\captionsetup{justification=centering}
		\caption{\textsc{Performance Comparisons for Individual Distortion Type on CVIQD Database. The Best Results are denoted in Bold.}}
		\label{table2}
		\scalebox{1.2}{
			\begin{tabular}{c|c|ccc|ccc|ccc}
				\toprule
				& Database  & \multicolumn{3}{c|}{JPEG} & \multicolumn{3}{c|}{AVC} & \multicolumn{3}{c}{HEVC} \\ \midrule
				Type & Methods & SROCC & PLCC & RMSE & SROCC & PLCC & RMSE & SROCC & PLCC & RMSE \\ \midrule
				\multirow{8}{*}{FR} & PSNR & 0.6982 & 0.8682 & 8.0429 & 0.5802 & 0.6141 & 10.552 & 0.5762 & 0.5982 & 9.4697 \\
				& S-PSNR \cite{yu2015framework} & 0.7172 & 0.8661 & 8.1008 & 0.6039 & 0.6307 & 10.3760 & 0.6150 & 0.6514 & 8.9585 \\
				& WS-PSNR \cite{sun2017weighted} & 0.6848 & 0.8572	& 8.3465 & 0.5521 & 0.5702 & 10.9841 & 0.5642 & 0.5884 & 9.5473 \\
				& CPP-PSNR \cite{zakharchenko2016quality} & 0.7059	& 0.8585 & 8.3109 & 0.5872 & 0.6137 & 10.5615 & 0.5689 & 0.6160 & 9.3009 \\
                & SSIM \cite{wang2004image} & 0.9582	& 0.9822 & 3.0468 & 0.9174 & 0.9303 & 4.9029 & 0.9452 & 0.9436 & 3.9097 \\
                & MS-SSIM \cite{wang2003multiscale} & 0.9047 & 0.9636 & 4.3355 & 0.7650 & 0.7960 & 8.0924 & 0.8011 & 0.8072 & 6.9693 \\
                & FSIM \cite{zhang2011fsim} & 0.9639	& 0.9839 & 2.8928 & 0.9439 & 0.9534 & 4.0327 & 0.9532 & 0.9617 & 3.2385 \\
				& DeepQA \cite{kim2017deep} & 0.9001 & 0.9526 & 4.9290 & 0.9375 & 0.9477 & 4.2683 & 0.9288	& 0.9221 & 4.5694 \\ \midrule
				\multirow{6}{*}{NR} & BRISQUE \cite{mittal2012no} & 0.9031 & 0.9464	& 5.2442 & 0.7714 & 0.7745 & 8.4573 & 0.7644 & 0.7548 & 7.7455 \\
				& BMPRI \cite{min2018blind} & 0.9562	& 0.9874 & 2.5597 & 0.6731 & 0.7161	& 9.3318 & 0.6715 & 0.6154 & 9.3071 \\
				& DB-CNN \cite{zhang2018blind} & 0.9576	& 0.9779 & 3.3862 & 0.9545 & 0.9564	& 3.9063 & 0.8693 & 0.8646 & 5.9335 \\
				& MC360IQA \cite{sun2019mc360iqa} & 0.9693 & 0.9698 & 3.9517	& 0.9569 & 0.9487 & 4.2281 & 0.9104	& 0.8976 & 5.2557 \\
				& VGCN \cite{xu2020blind} & \textbf{0.9759} & \textbf{0.9894} & \textbf{2.3590} & 0.9659 & 0.9719 & 3.1490 & 0.9432 & 0.9401 & 4.0257 \\
				& Proposed MFILGN & 0.9591 & 0.9862 & 2.7904 & \textbf{0.9683} & \textbf{0.9785} & \textbf{2.4998} & \textbf{0.9485} & \textbf{0.9581} & \textbf{3.3950} \\ \bottomrule
		\end{tabular}}
	\end{center}
\end{table*}

\begin{table*}[t]
	\begin{center}
		\captionsetup{justification=centering}
		\caption{\textsc{Performance Results for Various Weighting Strategies on OIQA and CVIQD Databases.}}
		\label{table3}
		\scalebox{1.2}{
			\begin{tabular}{c|ccc|ccc}
				\toprule
				Database  & \multicolumn{3}{c|}{OIQA} & \multicolumn{2}{c}{CVIQD} \\ \midrule
				Methods & SROCC & PLCC & RMSE & SROCC & PLCC & RMSE \\ \midrule
				Average Weighting & 0.9614 & 0.9695 & 0.5146 & 0.9670 & 0.9751 & 3.1036 \\
				Location Weighting & 0.9607 & 0.9688 & 0.5213 & 0.9665 & 0.9748 & 3.1212\\
				Content Weighting & 0.9598 & 0.9681 & 0.5252 & 0.9667	& 0.9749 & 3.1073 \\ \bottomrule
		\end{tabular}}
	\end{center}
\end{table*}

\textbf{1) Databases:} Two benchmark OIQA databases are utilized in the experiments, which consist of the OIQA \cite{duan2018perceptual} and CVIQD \cite{sun2017cviqd,sun2018large} databases.

\begin{itemize}
 \item \textbf{OIQA} comprises 16 pristine images and 320 distorted OIs degraded by 4 distortion types and 5 distortion levels. Among the distortion types, two kinds of compression artifacts are involved, namely JPEG and JPEG2000 compression. The remaining distortion types are Gaussian blur and Gaussian noise. The subjective quality ratings in the form of MOS are provided in the range [1, 10]. The higher MOS means better perceptual image quality.

\item \textbf{CVIQD} contains 528 compressed images derived from 16 original images. It adopts three popular image/video coding technologies, i.e. JPEG, H.264/AVC, and H.265/HEVC. Moreover, the MOS values in this database are normalized and rescaled to the range [0, 100].
\end{itemize}

\textbf{2) Measure Criteria:} To validate the effectiveness of our proposed MFILGN and compare with other state-of-the-arts, three commonly-used measure criteria \cite{video2003final} are employed as the following descriptions.

\begin{itemize}
\item \textbf{Spearman rank-order correlation coefficient (SROCC)} is computed by:

\begin{equation}\label{29}
SROCC=1-\frac{6\sum\limits_{i=1}^{N}{{{d}_{i}}^{2}}}{N({{N}^{2}}-1)},
\end{equation}
where $N$ is the number of image samples. ${d}_{i}$ indicates the rank difference between the subjective and objective evaluations for the $i-th$ image.

\item \textbf{Pearson linear correlation coefficient (PLCC)} is calculated as:

\begin{equation}\label{30}
PLCC=\frac{\sum\limits_{i=1}^{N}{({{s}_{i}}-{{\mu }_{{{s}_{i}}}})({{o}_{i}}-{{\mu }_{{{o}_{i}}}})}}{\sqrt{\sum\limits_{i=1}^{N}{({{s}_{i}}-{{\mu }_{{{s}_{i}}}})}*\sum\limits_{i=1}^{N}{({{o}_{i}}-{{\mu }_{{{o}_{i}}}})}}},
\end{equation}
where ${s}_{i}$ and ${o}_{i}$ denote the $i-th$ subjective and mapped objective quality values. ${{\mu }_{{{s}_{i}}}}$ and ${{\mu }_{{{o}_{i}}}}$ represent the corresponding mean values of ${s}_{i}$ and ${o}_{i}$, respectively.

\item \textbf{Root mean squared error (RMSE)} is defined by:

\begin{equation}\label{31}
RMSE=\sqrt{\frac{\sum\limits_{i=1}^{N}{{{({{s}_{i}}-{{o}_{i}})}^{2}}}}{N}}.
\end{equation}
\end{itemize}

\begin{table}[t]
	\begin{center}
		\captionsetup{justification=centering}
		\caption{\textsc{Performance Results for Different Viewport Numbers on OIQA and CVIQD Databases.}}
		\label{table4}
		\scalebox{0.95}{
			\begin{tabular}{c|ccc|ccc}
				\toprule
				Database  & \multicolumn{3}{c|}{OIQA} & \multicolumn{2}{c}{CVIQD} \\ \midrule
				Number & SROCC & PLCC & RMSE & SROCC & PLCC & RMSE \\ \midrule
				$6$  & 0.9608 & 0.9691 & 0.5155 & 0.9665 & 0.9746 & 3.1160 \\
				$20$ & 0.9614 & 0.9695 & 0.5146 & 0.9670 & 0.9751 & 3.1036 \\
				$80$ & 0.9616 & 0.9696 & 0.5134 & 0.9678 & 0.9758 & 3.0530 \\ \bottomrule
		\end{tabular}}
	\end{center}
\end{table}

In addition, each OIQA database is randomly divided into 80\% for training and the remaining 20\% for testing. We perform 1,000 iterations of cross validation on each database. The median SROCC, PLCC and RMSE performance values are then taken as the final measurement. Before calculating PLCC and RMSE for different objective quality assessment approaches, a five-parameter logistic nonlinear fitting function is used to map the predicted quality scores into a common scale as follows:

\begin{equation}\label{32}
g(x)={{\beta }_{1}}(\frac{1}{2}-\frac{1}{1+{{e}^{{{\beta }_{2}}(x-{{\beta }_{3}})}}})+{{\beta }_{4}}x+{{\beta }_{5}},
\end{equation}
where $\{{{\beta }_{i}}|i=1,2,...,5\}$ are five parameters to be fitted. $x$ and $g(x)$ denote the raw objective quality score and the regressed quality score after the nonlinear mapping process.

In addition, the above-mentioned three measure criteria can reflect different aspects of the performance for various IQA algorithms. Specifically, SROCC is generally used to measure prediction monotonicity, while PLCC and RMSE indicate prediction accuracy. Note that higher correlation coefficients and lower error mean better performance.

\subsection{Performance Comparison with Existing Objective Models}
In order to demonstrate the effectiveness of our proposed MFILGN model, we conduct extensive experiments compared to existing FR and NR objective image quality assessment algorithms. The FR models include conventional FR IQA as well as OIQA approaches (i.e. PSNR, SSIM \cite{wang2004image}, MS-SSIM \cite{wang2003multiscale}, FSIM \cite{zhang2011fsim}, S-PSNR \cite{yu2015framework}, WS-PSNR \cite{sun2017weighted} and CPP-PSNR \cite{zakharchenko2016quality}) and a deep learning-based FR IQA method (i.e. DeepQA \cite{kim2017deep}). Moreover, the NR models consist of traditional NR IQA approaches (i.e. BRISQUE \cite{mittal2012no} and BMPRI \cite{min2018blind}) and three deep learning-based NR IQA as well as OIQA methods (i.e. DB-CNN \cite{zhang2018blind}, MC360IQA \cite{sun2019mc360iqa} and VGCN \cite{xu2020blind}). Among these existing state-of-the-art objective FR image quality assessment models, the PSNR-related metrics including PSNR, S-PSNR, WS-PSNR and CPP-PSNR are the signal fidelity measurement, which compute the pixel differences between the reference and distorted images. Considering the characteristics of the HVS, the SSIM and its variants (i.e. MS-SSIM and FSIM) extract structural information from original and distorted images for perceptual image quality assessment. Besides, the DeepQA takes the human visual sensitivity into account in the deep learning framework. For NR methods, the BRISQUE is based on NSS features in the spatial domain and designed for conventional 2D IQA, and the BMPRI generates multiple pseudo reference images and exploits local binary pattern features for quality estimation. Additionally, the DB-CNN puts forward the bilinear pooling for predicting the perceptual image quality in the architecture of CNN. It is worth noting that the MC360IQA and VGCN are two deep learning-based methods specifically designed for 360-degree images. The MC360IQA utilizes 6 parallel sub-networks for viewport images, while the VGCN builds the graph convolution network based on different viewports.

\begin{table}[t]
	\begin{center}
		\captionsetup{justification=centering}
		\caption{\textsc{Performance Results for Different Image Decomposition Times by DHWT on OIQA and CVIQD databases.}}
		\label{table5}
		\scalebox{0.95}{
			\begin{tabular}{c|ccc|ccc}
				\toprule
				Database  & \multicolumn{3}{c|}{OIQA} & \multicolumn{2}{c}{CVIQD} \\ \midrule
				Time & SROCC & PLCC & RMSE & SROCC & PLCC & RMSE \\ \midrule
				$1$ & 0.9614 & 0.9695 & 0.5146 & 0.9670 & 0.9751 & 3.1036 \\
				$2$ & 0.9632 & 0.9716 & 0.4962 & 0.9671 & 0.9751 & 3.0992 \\
				$3$ & 0.9640 & 0.9715 & 0.4986 & 0.9675 & 0.9754 & 3.0945 \\ \bottomrule
		\end{tabular}}
	\end{center}
\end{table}

Table \ref{table1} shows the overall performance comparisons on OIQA \cite{duan2018perceptual} and CVIQD \cite{sun2017cviqd,sun2018large} databases. The best experimental results are highlighted in bold. The compared performance values are from \cite{xu2020blind}. For fair comparison, the performance values of traditional image quality assessment models are tested on the used testing data. Moreover, the learning-based models are trained on each specific 360-degree image quality database (i.e. OIQA database or CVIQD database) with randomly selected 80\% training data, and then tested on the remaining 20\% testing data. We can see that the PSNR-based metrics are inferior to other objective models considering the HVS properties. This is a common phenomenon because only signal errors are involved in the framework of PSNR-related models, which is far from the human perception. The 2D IQA models show unsatisfactory performance because they do not consider the specific characteristics of 360-degree images, such as the multiple viewports which are important for quality perception when browsing VR visual contents. Moreover, the deep learning-based models have advantages over traditional objective image quality assessment approaches for both FR and NR categories, especially the MC360IQA and VGCN which are two deep learning-based methods specifically developed for VR images. In addition, our proposed MFILGN achieves the best performance among the existing state-of-the-arts, including the deep learning-based OIQA and NSS-based IQA algorithms.

\begin{figure}[t]
  \centerline{\includegraphics[width=9cm]{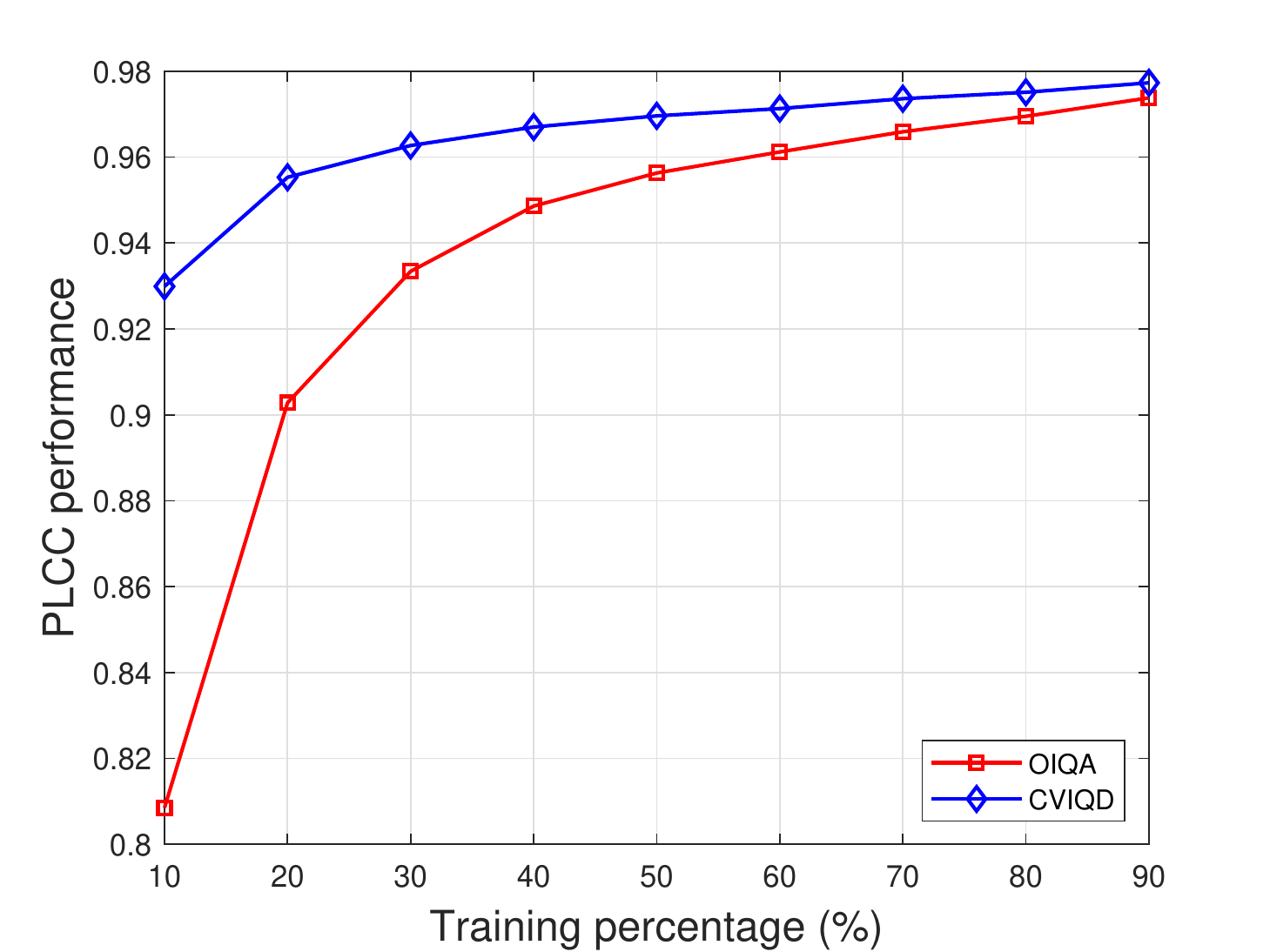}}
  \caption{The change of PLCC performance for our proposed MFILGN regarding to different training percentages on OIQA and CVIQD databases.}
  \centering
\label{fig:fig9}
\end{figure}

\begin{figure*}[t]
  \centerline{\includegraphics[width=16cm]{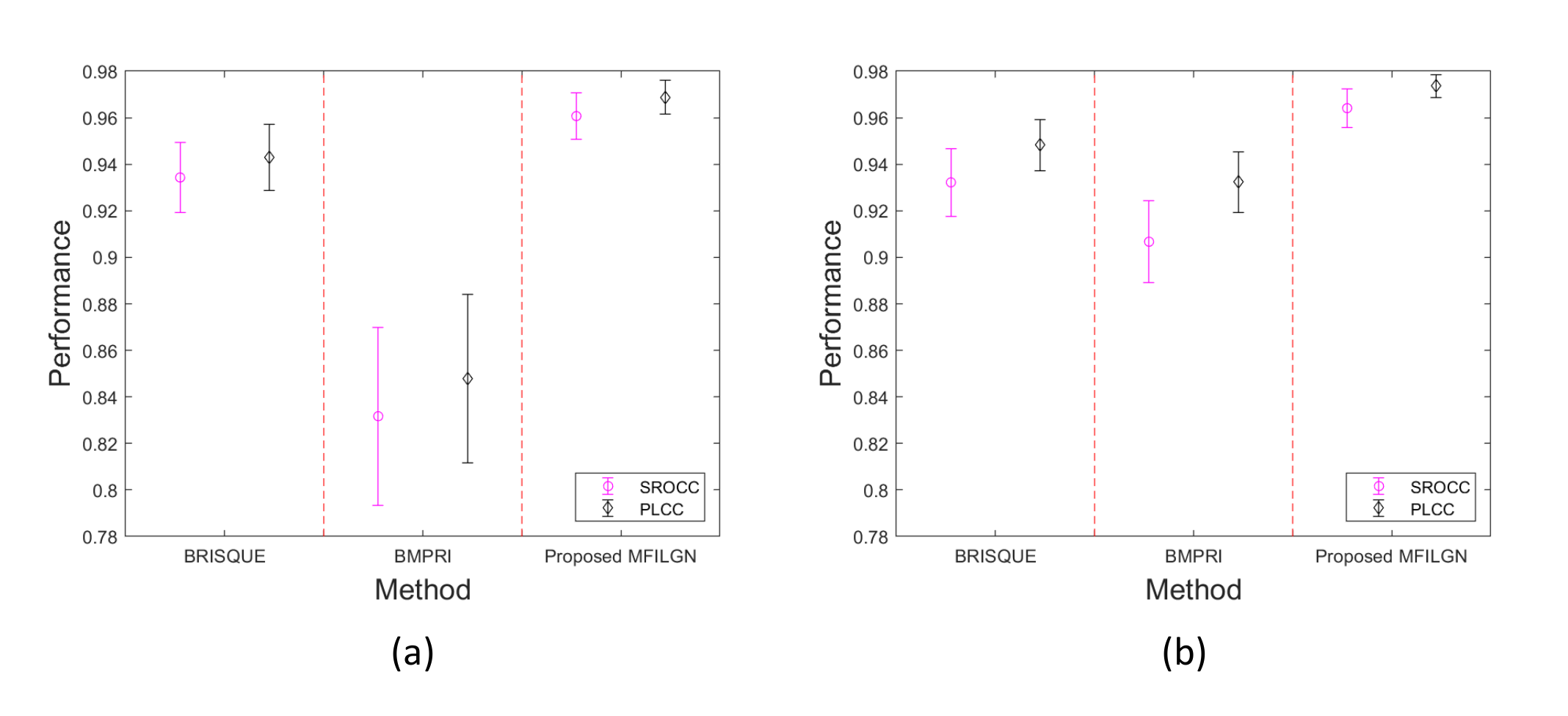}}
  \caption{Mean performance values and standard error bars for machine learning-based algorithms across 1,000 train-test trials. (a) Run on OIQA database; (b) Run on CVIQD database.}
  \centering
\label{fig:fig10}
\end{figure*}

\subsection{Performance Validity of Individual Distortion Type}
Since a variety of distortion types exist in current OIQA databases, we validate the performance regarding to each distortion type. As shown in TABLE \ref{table2}, the performance comparisons for individual distortion type are illustrated and the best results are denoted in bold. We can observe from this table that our proposed MFILGN outperforms other state-of-the-art models in terms of AVC and HEVC artifacts. For JPEG compression distortion, we can see that VGCN delivers the best performance. One possible explanation may be that the selected viewports of VGCN adopt graph modeling, which can better capture block effects caused by JPEG compression. In addition, even compared to deep learning-based methods specifically designed for 360-degree images (i.e. MC360IQA and VGCN), the proposed MFILGN can still achieve promising performance in the case of JPEG artifacts.

\subsection{Performance of Various Weighting Methods}
In the proposed MFILGN framework, we first extract local NSS features from multiple viewports. After the feature extraction, we then compute the average of these local NSS features to obtain the final local naturalness representations for each 360-degree image. Besides, except for the average weighting, there exist some other feature aggregation strategies, such as the location weighting and content weighting \cite{chen2020stereoscopic}. To be more specific, the location weighting strategy considers the statistics of eye-tracking data and uses the viewing probability to serve as the location weights for different viewport images. The content weighting method is based on the spatial information which reflects the spatial details of viewed regions.

We present the performance results for various weighting strategies on OIQA \cite{duan2018perceptual} and CVIQD \cite{sun2017cviqd,sun2018large} databases, as illustrated in TABLE III. We can find that our proposed MFILGN algorithm is insensitive to different weighting methods, which demonstrates the robustness of the proposed model. Therefore, for the sake of simplification, we choose the average weighting in the proposed MFILGN.

\subsection{Effects of Different Parameters}
Since viewport extraction is involved in our model, it is interesting to explore how the performance will be affected by different viewport numbers to be extracted. We test the performance results with respect to various viewport numbers on OIQA \cite{duan2018perceptual} and CVIQD \cite{sun2017cviqd,sun2018large} databases, as shown in TABLE \ref{table4}. Three cases are considered, which include the number equaling to 6, 20, and 80. As we can observe from this table, the performance of our proposed MFILGN gets better as the number of extracted viewports increases. However, the increasing viewport numbers inevitably could bring about more computational complexity. Thus, in order to find the balance between performance and computation, 20 viewports for each 360-degree image are utilized in our model, which is demonstrated in Fig. \ref{fig:fig8}.	

Furthermore, we change the DHWT decomposition times to see if using more image decompositions would lead to better performance. From the results listed in TABLE \ref{table5}, we can observe that by adding the number of DHWT layers, i.e. using more DHWT decomposition times, our model has a small performance boost. However, the increased layers of image decomposition could need to extract more features, which results in more computation time. In order to reduce the computation complexity, we choose one-time DHWT for measuring multi-frequency information, which can obtain the tradeoff between performance values and computation complexity.

In addition, we validate the change of PLCC performance of our proposed MFILGN regarding to different training percentages. As presented in Fig. \ref{fig:fig9}, in general, a large number of training data will bring about performance increase on both OIQA \cite{duan2018perceptual} and CVIQD \cite{sun2017cviqd,sun2018large} databases. In the proposed method, we choose 80\%-20\% for the training-testing split because this is a common practice for perceptual quality assessment in the literature \cite{chen2017blind,shi2019no,zhou2020tensor}. Besides, since the data distributions of OIQA and CVIQD databases are very different, the correlation performance for these two databases could be disparate. On one hand, even using 10\% training data, the MFILGN model can still deliver quite competitive performance results, especially for the CVIQD database, which further demonstrates the effectiveness of our proposed MFILGN method. On the other hand, one possible explanation for the performance differences between the two databases could be that the OIQA database seems more challenging compared to the CVIQD database.
\begin{table*}[t]
	\begin{center}
		\captionsetup{justification=centering}
		\caption{\textsc{Ablation Study on OIQA and CVIQD Databases. The Best Results are denoted in Bold.}}
		\label{table6}
		\scalebox{1.2}{
			\begin{tabular}{c|ccc|ccc}
				\toprule
				Database  & \multicolumn{3}{c|}{OIQA} & \multicolumn{2}{c}{CVIQD} \\ \midrule
				Methods & SROCC & PLCC & RMSE & SROCC & PLCC & RMSE \\ \midrule
				Multi-frequency information & 0.7734 & 0.7961	& 1.2666 & 0.7879 & 0.8285 & 7.8328 \\
				Global naturalness & 0.9260 & 0.9410 & 0.7106 & 0.9520 & 0.9599 & 3.9300 \\
				Local naturalness & 0.9460 & 0.9549 & 0.6195 & 0.9626 & 0.9723 & 3.2676 \\
                Local-global naturalness & 0.9495 & 0.9593 & 0.5915 & 0.9657 & 0.9739 & 3.1753 \\
				Proposed MFILGN & \textbf{0.9614} & \textbf{0.9695} & \textbf{0.5146} & \textbf{0.9670} & \textbf{0.9751} & \textbf{3.1036} \\ \bottomrule
		\end{tabular}}
	\end{center}
\end{table*}

\begin{table*}[t]
	\begin{center}
		\captionsetup{justification=centering}
		\caption{\textsc{Performance Comparisons for Cross Database Test by Training on CVIQD Database and Testing on OIQA Database. The Best Results are denoted in Bold.}}
		\label{table7}
		\scalebox{1.2}{
			\begin{tabular}{c|ccc|ccc|ccc}
				\toprule
				Database  & \multicolumn{3}{c|}{JPEG} & \multicolumn{3}{c|}{JPEG2000} & \multicolumn{3}{c}{ALL} \\ \midrule
				Methods & SROCC & PLCC & RMSE & SROCC & PLCC & RMSE & SROCC & PLCC & RMSE \\ \midrule
				MC360IQA \cite{sun2019mc360iqa} & 0.8412 & 0.8898 & 4.3950 & 0.6221 & 0.6211 & 5.1294 & 0.6981 & 0.7443 & 5.9184 \\
			    Proposed MFILGN & \textbf{0.8889} & \textbf{0.9027} & \textbf{0.9883} & \textbf{0.6781} & \textbf{0.7107} & \textbf{1.5545} & \textbf{0.7589} & \textbf{0.7885} & \textbf{1.3864} \\ \bottomrule
		\end{tabular}}
	\end{center}
\end{table*}

\subsection{Statistical Significance Analysis}
Since the compared BRISQUE \cite{mittal2012no}, BMPRI \cite{min2018blind}, and our proposed MFILGN are all based on the machine learning model called support vector regression, we repeat the process of database splitting for 1,000 times to compare the mean and standard deviation of performance values. We show the performance results in Fig. \ref{fig:fig10}, where the mean and standard deviations (std) of the SROCC and PLCC values across the 1,000 trials for three algorithms are illustrated. As can be seen in this figure, the proposed MFILGN method can achieve the higher mean value and smaller std compared to the others, which further suggests that MFILGN performs more precisely and consistently.

\subsection{Validity of Individual Proposed Quality Measure}
We explore the effectiveness of each proposed component in the MFILGN framework, namely multi-frequency information, local naturalness, global naturalness, and local-global naturalness. The performance values are provided in TABLE \ref{table6}. we can see that the local naturalness achieves the best performance, demonstrating the importance of viewports in evaluating the perceptual quality of 360-degree images. In addition, the multi-frequency information measurement can be used as a supplement to local and global naturalness features for further improving the performance of our proposed model. Especially for the OIQA database, by adding the multi-frequency information measurement, the SROCC performance improves from 0.9495 to 0.9614.

\subsection{Cross Database Test}
We validate the generalization capability of our proposed MFILGN model by cross database test, which is widely used in verification of model generalization ability. Considering that the CVIQD database has more compression distortion types than the OIQA database. Except for compression artifacts, the OIQA database contains Gaussian blur and Gaussian noise. Followed by \cite{sun2019mc360iqa}, we train objective quality assessment models on CVIQD database and then test JPEG and JP2000 compression for the OIQA database. The comparison results are shown in Table \ref{table7}. From this table, we can find that our MFILGN outperforms the state-of-the-art MC360IQA model \cite{sun2019mc360iqa} for both JPEG and JPEG2000 compression as well as the overall performance. It is also interesting to observe that testing on the JPEG compression distortion has a better performance compared to the JPEG2000 compression distortion. This is mainly because JPEG compression is the only common distortion t in both the two databases. To sum up, we can conclude that the proposed MFILGN method can achieve a good generalization capability.

\section{Conclusions}
In this paper, we present the Multi-Frequency Information and Local-Global Naturalness (MFILGN) scheme for no-reference quality assessment of omnidirectional images. The proposed MFILGN method is composed of two new measurements, including multi-frequency information measurement and local-global naturalness measurement. We design this model  by considering the HVS and the viewing process of 360-degree images. Specifically, based on the frequency-dependent property of visual cortex, we first exploit the multi-frequency channel decomposition to obtain both low-frequency and high-frequency subbands for 360-degree images. The entropy intensities of these subbands are then used to measure the multi-frequency information. Additionally, according to the viewing process, we adopt both local and global naturalness features from projected ERP maps and different viewports. The extracted features from our proposed two measuremenrs are fused together by the regression learning, which can predict the perceptual quality of omnidirectional images. We compare our proposed MFILGN with a number of state-of-the-art image quality assessment approaches on two publicly available 360-degree image quality databases. The experimental results demonstrate the superiority of our model.

We plan to develop a parametric model based on the proposed features and extend our method to omnidirectional video quality assessment. Furthermore, the optimization of VR processing systems based on our proposed blind quality assessment model is also promising in the future work.

\bibliographystyle{IEEEtran}
\bibliography{references}

%

%
%
%




\end{document}